\documentclass[letterpaper]{article} 
\usepackage{aaai24}  
\usepackage{times}  
\usepackage{helvet}  
\usepackage{courier}  
\usepackage[hyphens]{url}  
\usepackage{graphicx} 
\usepackage{natbib}  
\usepackage{caption} 
\usepackage{algorithm}
\usepackage{algorithmic}

\usepackage{newfloat}
\usepackage{listings}
\DeclareCaptionStyle{ruled}{labelfont=normalfont,labelsep=colon,strut=off} 
\floatstyle{ruled}
\newfloat{listing}{tb}{lst}{}
\floatname{listing}{Listing}
\title{SimFair: Physics-Guided Fairness-Aware Learning with Simulation Models}

\author {
    Zhihao Wang\textsuperscript{\rm 1},
    Yiqun Xie\textsuperscript{\rm 1}\thanks{Corresponding author.},
    Zhili Li\textsuperscript{\rm 1},
    Xiaowei Jia\textsuperscript{\rm 2},
    Zhe Jiang\textsuperscript{\rm 3},
    Aolin Jia\textsuperscript{\rm 1},
    Shuo Xu\textsuperscript{\rm 1}
}
\affiliations {
    \textsuperscript{\rm 1}University of Maryland\\
    \textsuperscript{\rm 2}University of Pittsburgh\\
    \textsuperscript{\rm 3}University of Florida\\
    \{zhwang1, xie, lizhili\}@umd.edu, xiaowei@pitt.edu, zhe.jiang@ufl.edu, \{aolin, shuoxu98\}@umd.edu
}

\usepackage{booktabs}
\usepackage{amsmath}
\usepackage{subfigure}
\usepackage{amsfonts,bm}
\usepackage{multirow}
\usepackage{braket}
\usepackage{mathtools}
\usepackage{enumerate}
\usepackage{enumitem}
\usepackage{empheq}
\usepackage{calc}
\usepackage{dsfont}
\usepackage{url}
\usepackage[flushmargin]{footmisc}
\usepackage{placeins}

\let\vec\textbf
\let\vv\relex

\newcommand{\vx}{\vec{X}}

\newcommand{\vy}{\vec{Y}}
\newcommand{\VX}{\vec{X}}
\newcommand{\VY}{\vec{Y}}
\newcommand{\vsx}{\vec{x}}
\newcommand{\vsy}{\vec{y}}
\newcommand{\vu}{\vec{u}}
\newcommand{\vv}{\vec{v}}
\newcommand{\ve}{\vec{e}}

\newcommand{\md}{\mathcal{D}}
\newcommand{\mf}{\mathcal{F}}
\newcommand{\ml}{\mathcal{L}}
\newcommand{\mm}{\mathcal{M}}

\newcommand{\vtheta}{\mathbf{\Theta}}

\newcommand{\hym}{\hat{\vy}^\mm}
\newcommand{\hy}{\hat{\vy}}

\DeclareMathOperator*{\argmin}{arg\,min}

\newtheorem{definition}{Definition}

\usepackage{fancyhdr}
\usepackage{float}
\begin{document}

\maketitle
\thispagestyle{fancy}

\pdfoutput =1

\begin{abstract}
Fairness-awareness has emerged as an essential building block for the responsible use of artificial intelligence in real applications.
In many cases, inequity in performance is due to the change in distribution over different regions.
While techniques have been developed to improve the transferability of fairness, a solution to the problem is not always feasible with no samples from the new regions,
which is a bottleneck for pure data-driven attempts.
Fortunately, physics-based mechanistic models have been studied for many problems with major social impacts.
We propose SimFair, a physics-guided fairness-aware learning framework, which bridges the data limitation by integrating physical-rule-based simulation and inverse modeling into the training design.
Using temperature prediction as an example, we demonstrate the effectiveness of the proposed SimFair in fairness preservation.
\end{abstract}

\section{Introduction}

As the use of artificial intelligence (AI) expands to more and more traditional domains, the bias in predictions made by AI has also raised broad concerns in recent years.
To facilitate the responsible use of AI, fairness-aware learning has emerged as an essential component in AI's deployment in societal applications.
In this study, we focus on learning-based mapping applications, where it is important to evaluate fairness over locations.
Such maps are often used to inform critical decision-making in major social sectors, such as food, energy, water, public safety, etc.

In these applications, especially at large scales, inequity in performance is often caused by changes in distribution over different regions \cite{xie2021statistically,goodchild2021replication}.
One of the major bottlenecks is the unavailability of ground truth data in test regions. 
With no labels from the test area (e.g., when applying models trained in one state to another), it is very difficult to know how to 
obtain fairness over new locations in the test area.
This is more challenging than transferring the overall prediction performance (e.g., measured by RMSE), which only needs to consider $ f: \vx \rightarrow \vy $ for the whole dataset. In the fairness-driven scenario, we also need to understand how the errors may vary over locations in a different region, which often does not follow the same pattern as the training region (e.g., the number of locations may vary; data distribution may vary).
Finally, the training and test areas often have completely different sets of locations, making the groups used in the fairness evaluation 
nonstationary as well.

In this paper, we use the temperature prediction problem as a concrete example. 
Air and surface temperatures are two key variables for estimating the Earth's energy budget, which connects to a diverse range of social applications, such as solar power, agriculture, climate change, global warming, ecosystem dynamics, and urban heat islands  \cite{kim1998feedbacks,peng2014afforestation,wang2023high,li2022impacts}. 
For example, temperature-related variables help estimate solar energy potential or predict the risks of floods or droughts at different locations. The results may affect resource allocation decisions such as subsidies, promotions, or insurance. 
Practically, satellite remote sensing is the only approach to measuring these variables at the spatial and temporal resolution needed for most applications \cite{liang2001optimization}. Due to the large volume of satellite data, machine learning methods have become increasingly popular choices in predicting temperature-related variables \cite{deo2017forecasting,wang2021method}.
However, fairness has yet to be considered.
Due to the social impact, it is important to ensure fairness among different places in the prediction map.

Given passive microwave and multi-spectral optical remote sensing imagery, the goal of the paper is to predict temperature while maintaining fairness among prediction performance over locations. In particular, we aim to improve 
the fairness of predictions in new test areas.

Recent studies have developed various approaches for fairness improvement.
On the data side, fairness-driven collection methods and filtering strategies were proposed to reduce bias caused by data issues such as imbalance \cite{jo2020lessons,yang2020towards,steed2021image}.
The methods are more suitable for domains where ground-truth data are reasonably easy to obtain. However, for most remote sensing problems, it is resource-intensive and time-consuming to collect new ground-truth samples (e.g., field surveys, sensor installation, and monitoring stations).
Many formulations explored decorrelating the feature learning process with sensitive attributes, which revealed information such as races and genders should not be discriminated in prediction \cite{zhang2021towards,alasadi2019toward,sweeney2020reducing}. For example, adversarial learning is a popular design choice in learning group-invariant features.
The use of regularization terms is another common approach to reduce bias risks, where fairness loss is used to penalize biased predictions \cite{yan2019fairst,serna2020sensitiveloss,zafar2017fairness}.
These methods, however, are not suitable for fair learning here between spatial regions as they require a fixed set of groups such as different genders, whereas the groups represented by locations vary between different regions.
There have also been studies for the time-series or online setups \cite{zhao2022adaptive,bickel2007discriminative,an2022transferring}. They aim to maintain fairness as new samples come in by sample reweighting, meta-learning, etc.
Similarly, these methods focus on fixed groups and are designed for dynamic changes in time series. They may also require additional ground-truth samples for finetuning.
Location-based fairness was also recently explored \cite{xie2022fairness, he2022sailing, he2023physics}, which reduced the statistical sensitivity in fairness evaluation for regression and classification tasks. However, it also requires training and test data from the same region.
Finally, all the above methods are purely data-driven, and their transferability is limited when no labels are available in a new region.

To address the limitations, we propose SimFair, a physics-guided fairness-aware learning approach, which uses simulations from mechanistic models to improve fairness in test regions. To the best of our knowledge, this is the first work that integrates physics-based simulation (mechanistic) models into fairness-aware learning. Our contributions include:
\begin{itemize}[leftmargin=*]
    \itemsep 2pt
    \item We present an inverse-modeling based design to integrate physics-based simulation models into the training process, which are often incompatible with the learning objectives in remote sensing problems.
    \item We propose a training strategy with dual-fairness consistency to improve fairness over new test locations.
    \item We incorporate physical-rule-based constraints to further improve the prediction performance.
    \item We integrate SimFair with different simulation models and real-world datasets for temperature prediction.
\end{itemize}

Through experiments, we demonstrate that the inverse modeling is robust, and SimFair can greatly improve fairness over new locations in test regions.

\section{Problem Definition}

\begin{definition}[Spatiotemporal (ST) domain]
Given a geographic space $S = \{s_1, \,s_2, \,...\}$ and a time-period $T = \{t_1, \,t_2, \,...\}$, a ST-domain $\md$ is a contiguous subspace in $S\times T$. For example, $\md$ can represent a contiguous geographic area (e.g., a county) over a month.
\end{definition}

\begin{definition}
[Location-based fairness measure]\label{def:fair}
It evaluates \textbf{prediction quality parity}, one of the standard definitions of fairness \cite{du2020fairness},
over a set of locations in a geographic region.
Denote $\mf$ as a prediction model; $\ml_p$ as the measure of prediction errors (e.g., RMSE); $\VX$ and $\VY$ as test features and labels, respectively; and $\vsx_i\in \VX$ and $\vsy_i \in \VY$ as features and labels for location $s_i\in S$, respectively.
Here the location-based fairness $\ml_f$ is defined as:
\begin{equation}\label{eq:fair}
    \ml_f = \frac{1}{|S|}\sum_{s_i\in S} \bigg|\ml_p(\mf(\vsx_i), \vsy_i) - \overline{\ml_p}(\mf(\VX), \VY) \bigg|
\end{equation}
$\ml_f$ evaluates the deviation of prediction performance from the global performance (i.e., a scalar obtained using entire test data $\VX$ and $\VY$). A smaller $\ml_f$ means the overall deviation is smaller, and thus the model is fairer over the locations.
\end{definition}

\paragraph{Formulation of location-based fair learning.} Given training samples $\VX$ and $\VY$ from a ST-domain $\md$, and test features $\VX'$ from a new ST-domain $\md'$, we aim to learn a (location-based) fairness-aware model from $\md$, which performs well in $\md'$ and, more importantly, offers fairer solution quality over locations in $\md'$.

A key characteristic of the problem is that the groups (i.e., locations $s\in S\in \md$) being considered are not prefixed and can be highly dynamic.
From one ST-domain to another, the locations being considered can be completely different (e.g., from one state to another).
This makes it difficult to connect the learning objectives from the training domain $\md$ to the target domain $\md'$.
Making the problem more challenging, only the features $\VX'$ are available from the new domain, and no label is available.
In essence, we need to build a fairness-aware model under distribution-shifts, different groups for fairness evaluation, and unknown labels.

\begin{figure*}
	\centering
        \vspace{5pt}
	\includegraphics[scale=0.424]{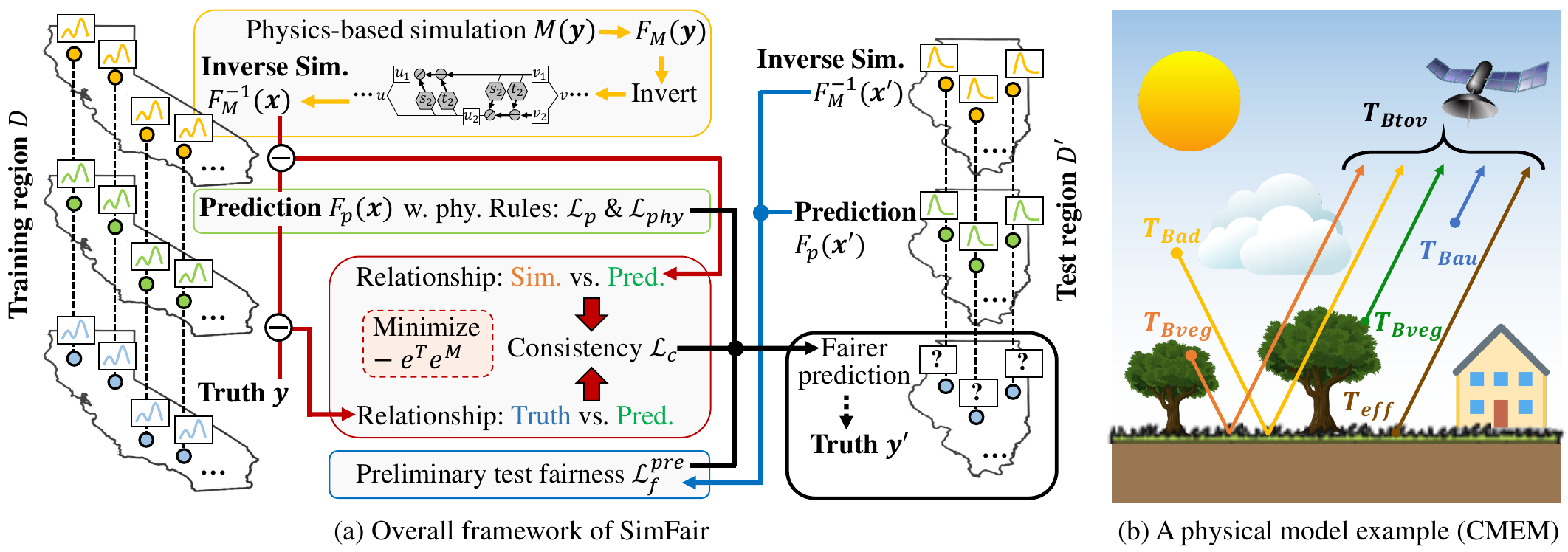}
	\caption{An illustration of simulation-model-guided fairness-aware learning.}
	\label{fig:overall}
\end{figure*}

\section{Method}
We propose SimFair, a physical-simulation-guided learning framework to improve the fairness-awareness of models for new ST-domains. 
To be concrete, we use temperature prediction as an example to illustrate the design.
In this section, we first provide brief overviews of two physics-based models we use, and then discuss the new SimFair framework.

\subsection{Physics-based Mechanistic Models}

\paragraph{Physics-Model 1 (PM1):}
The Community Microwave Emission Model (CMEM), as a subset of global operating systems at the European Centre, estimates low-frequency passive microwave brightness temperature (BT) \cite{kerr2010smos,wigneron2017modelling}. 
In the simulation process (Fig. \ref{fig:overall}(b)), CMEM computes the Top-of-Atmosphere (TOA)
BTs $T_{Btov,p,\theta}$ over vegetation layers for each polarisation direction $p$ and incidence angle $\theta$ by summing the soil effective temperature $T_{eff}$, vegetation temperature $T_{Bveg}$, and atmospheric components $T_{Bad}$ and $T_{Bau}$ (identical for high-altitude satellites). The overall physical process can be expressed as:
\begin{equation}\label{eq:cmem}
\begin{split}
    T_{Btov,p,\theta} = (1-\bm{r}_{r,p,\theta})T_{eff}\cdot\exp(-\tau_{veg,p,\theta})\\
    +T_{Bveg,p,\theta}(1+\bm{r}_{r,p,\theta}\cdot\exp(-\tau_{veg,p,\theta}))\\
    +T_{Bad,p,\theta}\cdot\bm{r}_{r,p,\theta}\cdot\exp(-2\tau_{veg,p,\theta})
\end{split}
\end{equation}
\noindent where $\bm{r}$ is soil surface reflectivity and $\tau$ is optical depth.

\paragraph{Physics-Model 2 (PM2):}
The MODerate resolution atmospheric TRANsmission (MODTRAN) model has been used worldwide to analyze, estimate, and predict the optical characteristics of the atmosphere based on the radiation transport physics \cite{berk2008band,berk2014modtran}. In remote sensing, TOA radiance, observed by satellites, is the mixed BT that is emitted, reflected, and transmitted by atmosphere and surface objects. MODTRAN simulation process is governed by: 
\begin{equation}\label{eq:MODTRAN}
    R_{i}(\theta) = (\varepsilon_i B_i(T_s)+(1-\varepsilon_i)R_{i\downarrow})\tau_i(\theta) + R_{i\uparrow}(\theta)
\end{equation}
\noindent where $R_{i}(\theta)$ is the TOA radiance captured by a certain range of wavelength (i.e., a satellite band) $i$ at a viewing zenith angle $\theta$; $R_{i\downarrow}$ and $R_{i\uparrow}$ represent the downward and upward atmospheric thermal radiance, respectively; $\varepsilon$ is land surface emissivity; $\tau$ is the atmospheric transmittance; and $B_i(T_s)$ denotes the Planck radiance at land surface temperature. 

\subsection{SimFair: Simulation-Enabled Fair Learning}\label{sec:method}

The overall framework of SimFair is illustrated in Fig. \ref{fig:overall}. 
Intuitively, we aim to learn the relationships between the data- and simulation-based predictions, and leverage these relationships to approximate fairness in a new test area.
SimFair has four components:
(1) inverse learning of the simulation models, which aligns the mechanistic model with a deep learning model; (2) preliminary test fairness, which weakly estimates fairness in the test region using simulations but by itself is insufficient to improve fairness;
(3) a dual-fairness consistency, which tries to minimize the gap between data- and simulation-based fairness; and (4) physical rules, which are used as soft constraints to improve generalizability.

\subsubsection{Inverse Modeling for Learning.}
In physics-based modeling, the processes are not necessarily derived from a direction that aligns with the one we use in prediction tasks. For example, in temperature simulation for passive remote sensing (PM1), the real physical process starts from the air or surface temperature, where radiance travels through the air -- being absorbed, reflected/deflected, emitted, or transmitted by vegetation, built-ups, atmospheric particles, etc., -- and finally reaches the spectral sensor from the satellites and recorded as signal values. This process can be described as $\vx = \mm(\vy)$ where $\vx$ represents satellite signals, $\vy$ is the temperature, and $\mm$ is the mechanistic model. However, in real-world applications, it often goes in the opposite direction, where users predict the temperature (i.e., $\vy$) using satellite readings $\vx$. Having consistent directions is important for the use of simulation models in guiding data-driven approaches, because for each observation $\vsx_i$ 
we need to know the corresponding simulated value $\vsy_i = \mm^{-1}(\vsx_i)$ (e.g., temperature) to extract useful information.
Unfortunately, it is often very difficult to directly find the inverse of a mechanistic model due to the complexity of the physical process. For example, there are no known inversions of the mechanistic models PM1 and PM2 used here.

To address this issue, we first use bijector-based invertible networks \cite{kobyzev2020normalizing,kingma2016improved,dinh2016density} to approximate the inverses of physics-based models; the structures were often used in normalizing flows for the estimation of complex statistical distributions and random sampling. 
While the direction can also be reversed in vanilla neural networks by swapping $\vx$ and $\vy$, we use the invertible design for three major reasons:
\begin{itemize}[leftmargin=*]
    \item In physical processes $\vx = \mm(\vy)$, many physics constraints can only be used on the variables in $\vx$ (the constraints are built into the loss later in a neural network).
    There is no problem if we train an invertible network using direction $\vx = \mf(\vy)$ and then inverse it. However, if we simply use a data swap $\vy = \mf(\vx)$, we can no longer apply the constraints, as $\vx$ are fixed inputs for training instead of outputs.
    \item The invertible structure naturally provides extra regularization, as the learned weights need to work simultaneously for both directions, improving prediction quality during the test (evaluated later in experiments).
    \item When $\vsx_i$ and $\vsy_i$ have different lengths, the invertible structure can be naturally extended with normalizing flow to quantify the uncertainty from fewer to more variables.
\end{itemize}

In the application context, we denote $\vx$ as satellite signals, $\vy$ as the prediction target (e.g., temperature), $\mm$ as the mechanistic model, $\mf_\mm(\cdot; \vtheta)$ as an invertible neural network, $\mf_\mm^{-1}(\cdot; \vtheta)$ as its inverse, and $\ml$ as a loss function (e.g., RMSE). The inverse approximation is given by:
\begin{equation}\label{eq:inn}
\mf_\mm^{-1}(\vx; \vtheta^*=\argmin_\vtheta 
\ml(\mm(\vy), \mf_\mm(\vy, \vtheta))\big)
\end{equation}
The bijector-based invertible layers present a great fit for the inverse approximation because (1) while complex, a physics-based mechanistic model describes a single function, i.e., all simulated labels $\mm(\vx)$ follow the same distribution $P(\mm(\vx)\,|\,\vx)$. 
Thus, $\mf_\mm(\cdot; \vtheta)$ can effectively approximate $\mm(\vx)$ given the capability of deep neural networks to universally approximate continuous functions. 
(2) 
Bijectors use mathematically exact inversion, enabling us to create a highly accurate approximation of the inverse of $\mm(\vx)$.
Specifically, we use the following formulation \cite{dinh2016density}:
\begin{equation}
    \begin{split}
    &\begin{cases}
    \vv_{1} &= \vu_{1} \odot \exp\big(s_2(\vu_{2})\big) + t_2(\vu_{2})\\
    \vv_{2} &= \vu_{2} \odot \exp\big(s_1(\vv_{1})\big) + t_1(\vv_{1})\\
    \end{cases}\\
    &\Longleftrightarrow\begin{cases}
    \vu_{2} &= \big(\vv_{2} - t_1(\vv_{1})\big) \odot \exp\big(-s_1(\vv_{1})\big)\\
    \vu_{1} &= \big(\vv_{1} - t_2(\vu_{2})\big) \odot \exp\big(-s_2(\vu_{2}) \big)\\
    \end{cases}
    \end{split}
\end{equation}
\noindent where $\vu$ and $\vv$ are the input and output of an invertible layer, respectively; $\vu = [\vu_1, \vu_2]$ and $\vv = [\vv_1, \vv_2]$; $s_1(\cdot)$, $t_1(\cdot)$, $s_2(\cdot)$, and $t_2(\cdot)$ are learnable functions.
Note that both the input and output have the same length, which is a property of the bijector needed to make inversions.

Using a chain of bijectors as network layers, we construct the invertible network $\mf_\mm$ to approximate the inverse of $\mm$. As the parameters we are interested in are a subset of those in the complete mechanistic models, we select the most related ones to define the original input and final output of the chain of bijectors $\mf_\mm$, which also allows us to make their lengths equivalent. Through experiments, we found that this led to approximations with higher precision for both directions of $\mf_\mm$ (i.e., $\hat{\vx} = \mf_\mm(\vy)$ and $\hat{\vy} = \mf_\mm^{-1}(\vx)$), compared to the formulations where $\vx$ and $\vy$ had different lengths (a random vector $\vec{z}$ needs to be added to the shorter one in this case, which is also a more flexible option).

\subsubsection{Preliminary Fairness on Test Region.}
This is the first component of SimFair. As shown in Fig. \ref{fig:overall}, we aim to approximate the fairness between data samples at different locations in the test region $\md'$ using relationships between simulation- and learning-based predictions. Here the results from the inverse simulation model $\hat{\vy}^\mm$, obtained through the invertible network $\mf_\mm^{-1}(\vx)$, 
provide us a preliminary peek into the labels $\vy$ from the test region $\md'$.

Here we need to emphasize that as there is no guarantee about the distances between $\hat{\vy}^\mm$ and real labels $\vy$ (or the variance of the distances), fairness scores evaluated using Eq. (\ref{eq:fair}) with $\hat{\vy}^\mm$ as the truth are not directly representative of the true fairness.
Thus, the goal of this part is only to create a "preliminary fairness" as a preparation step for the dual-fairness consistency module in the next section, where new designs will be used to bridge the gap.

With that clarified, the preliminary fairness loss on $\md'$ is:
\begin{equation}\label{eq:fair_sim}
\small
    \ml^{pre}_f = \frac{1}{|D'|}\sum_{\vsx_i\in D'} \bigg|\ml_p(\mf_p(\vsx_i), \hat{\vsy}^\mm_i) - \overline{\ml_p}(\mf_p(\vx), \hat{\vy}^\mm) \bigg|
\end{equation}
\noindent where $\vsx_i$ represents the data point at location $s_i$, $\hat{\vy}^\mm = \mf_\mm^{-1}(\vx)$, $\ml_p$ is the prediction loss, and $\mf_p$ is the prediction model, which is used to predict the real values rather than approximate the simulation model $\mm$.

\subsubsection{Dual-Fairness Consistency.}
The dual-fairness consistency module aims to reduce the gap between the preliminary fairness loss $\ml_f^{pre}$ and the real fairness loss 
$\ml_f^{real}$ 
(not evaluable in training) for the test region $\md'$. It achieves this by learning and governing the triplet relationships among the following in the training data: 
\begin{itemize}
    \item Physical simulations $\hat{\vy}^\mm$ (inversely approximated);
    \item Deep neural network predictions $\hat{\vy}$; and
    \item True labels $\vy$.
\end{itemize}

\begin{figure}
	\centering
        \vspace{5pt}
	\includegraphics[scale=0.42]{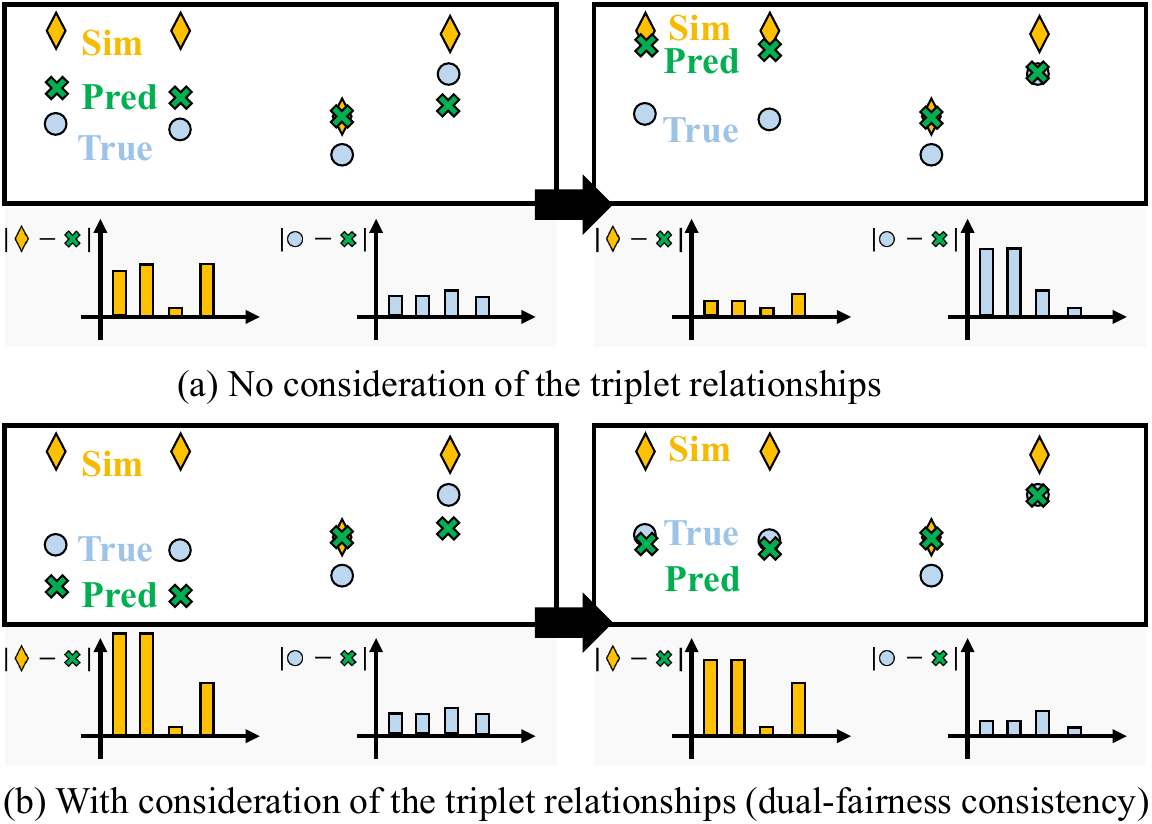}
	\caption{Illustrative example of dual-fairness consistency.}
	\label{fig:dual}
\end{figure}

While we do not know the relationships between the simulation results $\hat{\vy}^\mm$ and true labels $\vy$ in the test region $\md'$, we can find a solution $\mf_p$ whose predictions $\hat{\vy}$ have a similar relationship with simulated-based $\hat{\vy}^\mm$ and true $\vy$.
Specifically, for the triplet relationship, our desired property is:

\begin{definition}[Dual-fairness consistency]
Denote $\ve = \vy - \hat{\vy}$ and $\ve_\mm = \hat{\vy}^\mm - \hat{\vy}$, which represent the differences between the true labels and the predictions, and the simulated labels and the predictions, respectively. 
Dual-fairness refers to the fairness evaluation defined using true labels (Eq. (\ref{eq:fair})) and simulation labels (Eq. (\ref{eq:fair_sim}); here for training data in $\md$), respectively.
To make the two fairness results more consistent, we aim to align the direction of the predicted labels $\hat{\vy}$ with respect to the true labels $\vy$ and simulation labels $\hat{\vy}^\mm$:
\begin{equation}
\begin{cases}
\ve_i \geq 0, &\text{if } (\ve_\mm)_i\geq 0\text{;}\\
\ve_i < 0, &\text{otherwise.}
\end{cases}    
\end{equation}
\noindent where $i$ denotes the $i^{th}$ data point.
\end{definition}

Fig. \ref{fig:dual} shows the high-level idea with an illustrative example. As we can see, the relationships, i.e., $\ve$ and $\ve_\mm$, are often not aligned in Fig. \ref{fig:dual} (a), which does not consider the consistency. As a result, improving fairness w.r.t. the simulation results leads to a less fair result. In contrast, with the consistency, the fairness improvement w.r.t. simulation data is more likely to lead to improvements w.r.t. the true data.
Intuitively, when the directions represented by $\ve_i$ and $(\ve_\mm)_i$ are aligned, reducing the distance between a prediction $\hy_i$ and a simulation label $\hym_i$ will accordingly reduce the distance between $\hy_i$ and $\vsy_i$. Note that since both $\vy$ and $\hym$ are fixed inputs at this stage and they are not identical, it is impossible to make $\ve = \ve_\mm$. Instead, our focus here is to promote a solution $\mf_p$ if it maintains similar directional relationships between $\ve$ and $\ve_\mm$. This is important for reducing the fairness loss, as we are trying to re-balance the prediction losses among points at different locations while keeping a similar global prediction loss (Eq. (\ref{eq:fair}) or (\ref{eq:fair_sim})). In other words, the fairness loss moves a prediction closer to the true label if the loss is worse than the global mean loss, and farther otherwise. Based on the dual-fairness consistency, we have the consistency loss as:
\begin{equation}
\begin{split}
    &\ml_c = - \ve^T\ve_\mm\\
    & = -\sum_{(\vsx_i,\vsy_i) \in \md} \big(\vsy_i - \mf_p(\vsx_i)\big)
     \cdot \big(\mf_\mm^{-1}(\vx) - \mf_p(\vsx_i)\big)
\end{split}
\end{equation}
$\ml_c$ allows gradients based on the preliminary test fairness loss $\ml^{pre}_f$ to be more reflective of the true fairness loss.

\subsubsection{Improvements with Physics-guided Predictions.}
We incorporate physical constraints from the mechanistic models as part of the loss functions to reduce the overfitting of the prediction model $\mf_p$ \cite{jia2021physics,chen2023physics}, which accordingly makes it generalize better to the test region $\md'$.
As physical models rely on different assumptions, we use two different constraints for the two physical models (i.e., PM-1 and PM-2). The physical rule used for PM-1 is the Rayleigh-Jeans Law of radiation, which states the radiance emitted by a gray body (e.g., trees, rocks) is less than a black body with unity emissivity.  
The loss $\ml^{PM1}_{phy}$ is then:
\begin{equation}
\textbf{1}^T
    [\max\big(0,\vx - \bm{\varepsilon}\otimes\mf_p(\vx)\big)
    + \min\big(0,\vx - \bm{\eta}\otimes\mf_p(\vx)\big)]
\end{equation}
\noindent 
where $\bm{\varepsilon} = \min(\textbf{1},\mf_\mm(\vsy_i)\otimes(\mf_\mm^{-1}(\vsx_i)))^{-1}$,
$\bm{\eta} = \max(\textbf{0},$ $\mf_\mm(\vsy_i)\otimes(\mf_\mm^{-1}(\vsx_i)))^{-1}$, and $\otimes$ is the Hadamard product.

For PM-2, the model output (temperature) is bounded by a well-known principle, the surface energy balance equation. Specifically, in the solar-earth energy exchange system, the overall energy is balanced by solar downward shortwave $R_{S\downarrow}$ and longwave $R_{L\downarrow}$, surface upward shortwave $R_{S\uparrow}$ and longwave $R_{L\uparrow}$, and net radiance $R_N$. The balance of energy at the surface is also related and can be expressed as the combination of upward surface sensible heat flux $H_S$, upward surface latent heat flux $H_L$, and downward ground heat flux $H_G$. This leads to the following loss $\ml^{PM2}_{phy}$:
\begin{equation}\label{eq:energy}
    -\varepsilon\sigma \mf_p(\vx)^4+R_{S\downarrow}-R_{S\uparrow}+\varepsilon R_{L\downarrow}- (H_S + H_L + H_G)
\end{equation}
\noindent where 
$\varepsilon$ is the surface emissivity, and $\sigma$ is the Stefan-Boltzmann contant. 
Finally, the overall loss 
is 
$\ml = \ml_p + \ml^{pre}_f + \ml_c + \ml_{phy}$,
where $\ml_{phy}$ is selected based on the physical model used (e.g., PM-1), $\ml_p$ we use in the paper is mean squared loss $\ml_p = ||\mf_p(\vx) - \vy||^2_2/|\md|$, 
and all losses are normalized based on the number of samples.

\begin{table*}
\vspace{5pt}
\footnotesize
\centering
\begin{tabular}{|c|l||ccc||ccc||ccc|}
\hline
& & \multicolumn{3}{c||}{Train-Test: West-East}
& \multicolumn{3}{c||}{Train-Test: East-West}
& \multicolumn{3}{c|}{Train-Test: East-Alaska}
\\\hline
& Model     & RMSE & Corr. & Fairness              & RMSE  & Corr. & Fairness              & RMSE  & Corr. & Fairness              \\
\hline
\multirow{8}{*}{\rotatebox[origin=c]{90}{\footnotesize FNN}}
& BaseNet  & 6.75 & 0.83  & 4.49 (±0.81)          & 27.56 & 0.34  & 13.65 (±2.48)         & 52.03 & 0.19  & 44.42 (±3.37)         \\
& Sim       & 6.45 & 0.86  & 4.69 (±0.82)         & 20.33 & 0.44  & 11.93 (±2.19)         & 43.74 & 0.29  & 38.84 (±5.69)         \\
& SimPhy    & 7.19 & 0.84  & 5.56 (±0.4)          & 17.78 & 0.48  & 10.79 (±1.98)          & 45.52 & 0.29  & 38.84 (±3.4)         \\
& RegFair       & 7.22 & 0.8 & 4.97 (±0.69)           & 25.35 & 0.37 & 12.36 (±2.42)         & 38.5 & 0.06 & 29.73 (±6.78)         \\
& Self-Reg      &  6.35 & 0.84 & 4.27 (±0.7)           & 31.97 & 0.31 & 16.48 (±2.42)          & 38.01 & 0.06 & 28.95 (±4.15)        \\
\cline{2-11}
& SimFair   & 3.07 & 0.97  &2.04 (±0.19) & 3.11  & 0.96  & \textbf{1.94 (±0.03)} & 6.23  & 0.84  & \textbf{4.25 (±0.78)} \\
& SimFair-P & 2.88 & 0.97  & \textbf{1.89 (±0.06)} & 3.13  & 0.96  & 1.96 (±0.05) & 6.29  & 0.81  & 4.45 (±0.51)
\\
\hline
\multirow{8}{*}{\rotatebox[origin=c]{90}{\footnotesize LSTM}}
& BaseNet  & 4.22	& 0.93 & 2.66	(±0.14)         & 4.02&0.97&2.45	(±0.16)        & 11.93&0.8&5.14	(±0.31)        \\
& Sim       & 3.89 & 0.95 & 2.43	(±0.15)        & 3.3&0.97&2.21	(±0.40)       & 13.32&0.85&5.25	(±0.49)        \\
& SimPhy    & 4.46 & 0.95 & 2.69	(±0.17)        & 3.23&0.97&2.04	(±0.17)          & 12.27&0.88&4.82	(±0.28)      \\
& RegFair       & 4.17 & 0.94 & 2.66	(±0.22)        & 4.03&0.96&2.59	(±0.58)        & 12.16&0.81&5.03(±0.4)   \\
& Self-Reg      &  4.10 & 0.94 & 2.57	(±0.26)        & 3.85&0.96&2.41	(±0.16)        & 11.24&0.84&4.68(±0.41)   \\
\cline{2-11}
& SimFair   & 3.46 & 0.96 & 2.21	(±0.11) & 3.22&0.98&\textbf{1.91 (±0.11)} & 11.05&0.86&4.55(±0.27) \\
& SimFair-P & 3.35 & 0.96 & \textbf{2.12(±0.11)} & 3.24&0.97&1.99	(±0.17) & 10.52&0.89&\textbf{4.15(±0.23)}
\\
\hline
\end{tabular}
\caption{AT1: Fairness results on temperature prediction (split by geographic regions in Fig. \ref{fig:data}(a)). 
}
\label{tab:ew}
\end{table*}

\subsection{Deep Networks}
We implemented SimFair using two types of networks: (1) a fully-connected neural network, FNN, which uses observed signals from satellite snapshots to make predictions;
and (2) a long-short-term-memory (LSTM) model that uses a time-series-based input. 
Our invertible network uses a chain of 7 bijector layers. We use root-mean-squared-errors (RMSE) as the loss function and the Adam optimizer with an initial learning rate of {$10^{-2}$}.
More details are in the Appendix.

\begin{figure}
\centering
\vspace{5pt}
\includegraphics[width=0.95\columnwidth]{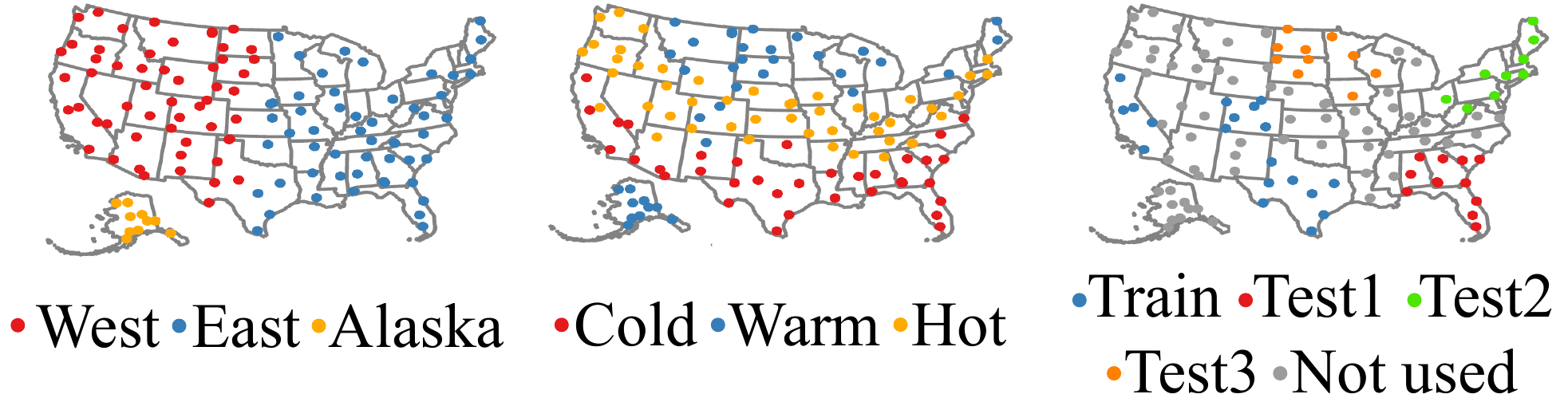}
\caption{Spatial distributions of training and testing data.}
\label{fig:data}
\end{figure}

\section{Experiments}

\subsection{In-Situ and Remote Sensing Datasets}

We use three real datasets for evaluation: AT1, AT2 and LST (detailed in the following paragraphs).
As AT1 contains the largest number of high-quality stations (122), we use it to evaluate the models' ability to promote fairness in test regions, that contain different locations from the training region.
Additionally, we include two smaller datasets AT2 and LST,
which are used to evaluate if fairness learned among the 7 locations during a period can be transferred to a new period (i.e., the same set of locations over different periods).

\paragraph{AT1: USCRN-CMEM air temperature data.}
We collected the ground truth from the entire USCRN stations (200+) in 2014. All station measurements
were carefully examined, and only the dates with high-quality measurements were used. Satellite observations were collected from the AMSR2 satellite with two observations per day. 
CMEM (PM1) model inputs were collected from ERA5 hourly dataset, including soil temperature, volumetric soil water layer, etc. Other surface and atmospheric datasets were simulated using ecoClimate.
We used three types of space partitionings to create train-test splits (Fig. \ref{fig:data}) with different geographic regions, temperature zones, and random local states.

\paragraph{AT2: SURFRAD-CMEM air temperature data.}
Different from AT1, 
the ground truth in AT2 was collected from a well-known and high-quality network, SURFRAD, which measured surface conditions and energy at minute scales.
As discussed, we separated the data using two temporal splits: (1) first 8 months as training and last 4
months as test; and (2) first 4 months as training and last
8 months as test.

\paragraph{LST: SURFRAD-MODTRAN land surface temperature data.}
We collected the surface temperature and four radiance measurements in SURFRAD stations from 2013 to 2020. Satellite observations were collected from Landsat images.
MODTRAN (PM2) inputs were collected from NCEP Reanalysis and ASTER Global Emissivity products. We split train-test by: (1) first 5 years as training and last 3 as test; and (2) first 3 years as train and last 5 as test.

\subsection{Results and Analysis}

For the three datasets (AT1, AT2, LST), we evaluate the following methods with the same fairness definition in Eq. (\ref{eq:fair}):
\begin{itemize}[leftmargin=*]
    \itemsep 0pt
    \item \textbf{BaseNet:} This is the baseline neural network, i.e., FNN or LSTM, without additional fairness consideration.
    \item \textbf{Sim:} BaseNet that uses the physics-based simulation data in pre-training \cite{jia2021physics,li2022estimating}. This provides a more generalizable initialization of the model.
    \item \textbf{SimPhy:} This approach uses both simulation-based pre-training as well as physical constraints in loss design to regularize the training and improve generalizability to test samples from different regions \cite{willard2020integrating}.
    \item \textbf{RegFair:} This is the regularization-based fairness-aware learning \cite{serna2020sensitiveloss,yan2019fairst}, which includes additional fairness-related loss to learn a fairer model on the training dataset.
    \item \textbf{Self-Reg:} A self-training based fair learning framework, which uses predicted labels on the test data to create a pseudo-fairness-loss to adapt to the test area. The predicted labels are dynamically updated during training.
    \item \textbf{SimFair:} Proposed approach (no physical constraints).
    \item \textbf{SimFair-P:} Complete version with physical constraints.
\end{itemize}

\begin{table*}[ht]
\vspace{5pt}
\footnotesize
\centering
\begin{tabular}{|l||ccc||ccc||ccc|}
\hline
& \multicolumn{3}{c||}{Train-Test: Hot-Cold}
& \multicolumn{3}{c||}{Train-Test: Cold-Hot}
& \multicolumn{3}{c|}{Train-Test: Hot-Warm}
\\\hline
Model     & RMSE & Corr. & Fairness              & RMSE  & Corr. & Fairness              & RMSE  & Corr. & Fairness              \\
\hline
BaseNet   & 19.7 & 0.56 & 12.36(±1.78) & 35.95 & 0.25 & 21.87(±2.92) & 17.5 & 0.61 & 12.1(±0.88) \\
Sim       & 20.26 & 0.5 & 13.71(±1.86) & 35.69 & 0.18 & 22.19(±1.43) & 15.86 & 0.58 & 11.54(±1.19) \\
SimPhy    & 17.84 & 0.59 & 11.28(±1.63) & 35.37 & 0.2 & 22.32(±3.16) & 16.35 & 0.59 & 11.83(±1.) \\
RegFair       & 19.42 & 0.57 & 11.9(±0.29) & 35.48 & 0.23 & 21.69(±0.8) & 16.95 & 0.63 & 11.86(±1.32) \\
Self-Reg      & 19.32 & 0.58 & 12.(±0.55) & 36.11 & 0.24 & 21.81(±0.61) & 16.27 & 0.65 & 11.03(±0.54) \\
\hline
SimFair   & 11.97 & 0.88 & 4.8(±1.24) & 9.25 & 0.78 & 3.61(±0.56) & 5.77 & 0.9 & 3.62(±0.44) \\
SimFair-P & 12.37 & 0.91 & \textbf{4.43(±0.78)} & 9.42 & 0.78 & \textbf{3.46(±0.22)} & 5.62 & 0.89 & \textbf{3.56(±0.21})
\\
\hline
\end{tabular}
\caption{AT1: Fairness results on temperature prediction (split by temperature zones in Fig. \ref{fig:data}(b)). 
}
\label{tab:t}
\end{table*}

\begin{table*}
\footnotesize
\centering
\begin{tabular}{|l||ccc||ccc||ccc|}
\hline
& \multicolumn{3}{c||}{Train-Test: Train-Test1}
& \multicolumn{3}{c||}{Train-Test: Train-Test2}
& \multicolumn{3}{c|}{Train-Test: Train-Test3}
\\\hline
Model     & RMSE & Corr. & Fairness              & RMSE  & Corr. & Fairness              & RMSE  & Corr. & Fairness              \\
\hline
BaseNet   & 24.02 & 0.13 & 14.22(±0.98) & 28.58 & 0.51 & 19.48(±3.16) & 27.87 & 0.56 & 15.68(±0.83) \\
Sim       & 22.72 & 0.25 & 14.24(±1.63) & 29.71 & 0.41 & 19.97(±2.98) & 22.18 & 0.54 & 11.59(±2.59) \\
SimPhy    & 22.86 & 0.25 & 14.02(±2.59) & 28.21 & 0.43 & 19.35(±2.94) & 21.64 & 0.60 & 10.88(±0.98) \\
RegFair       & 23.71 & 0.14 & 14.22(±0.62) & 30.66 & 0.47 & 20.87(±2.83) & 26.77 & 0.57 & 15.30(±2.26) \\
Self-Reg      & 25.58 & 0.13 & 15.57(±1.65) & 28.70 & 0.49 & 19.26(±2.41) & 26.55 & 0.56 & 14.79(±1.52) \\
\hline
SimFair   & 8.42 & 0.82 & 5.37(±0.52) & 6.55 & 0.90 & 3.52(±0.30) & 10.01 & 0.92 & 5.06(±0.38) \\
SimFair-P & 7.52 & 0.86 & \textbf{4.86(±0.31)} & 5.94 & 0.90 & \textbf{3.26(±0.28)} & 9.64 & 0.91 & \textbf{4.90(±0.49)}
\\
\hline
\end{tabular}
\caption{AT1: Fairness results on temperature prediction (split by random state groups in Fig. \ref{fig:data}(c)). 
}
\label{tab:state}
\end{table*}

\begin{figure} 
\centering
\subfigure[$X,Y$ data swap approx.]{ \label{fig:a}{}
\includegraphics[width=0.38\columnwidth]{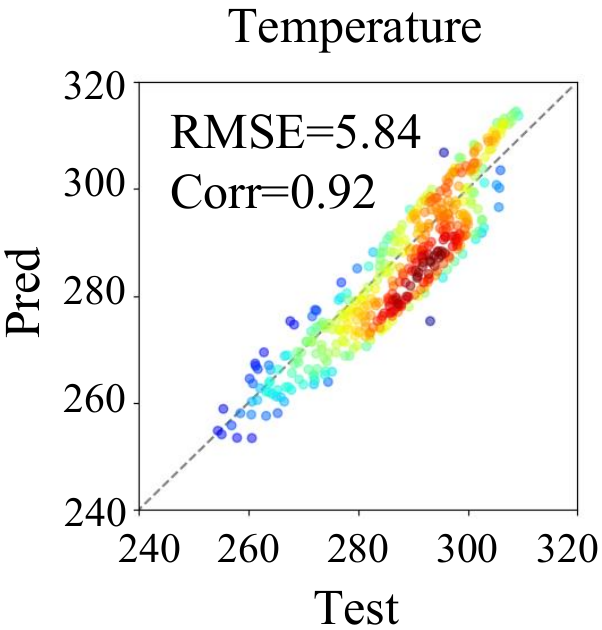}
}
\subfigure[$\mf^{-1}_\mm$: inverse approx.]{ \label{fig:b}{}
\includegraphics[width=0.38\columnwidth]{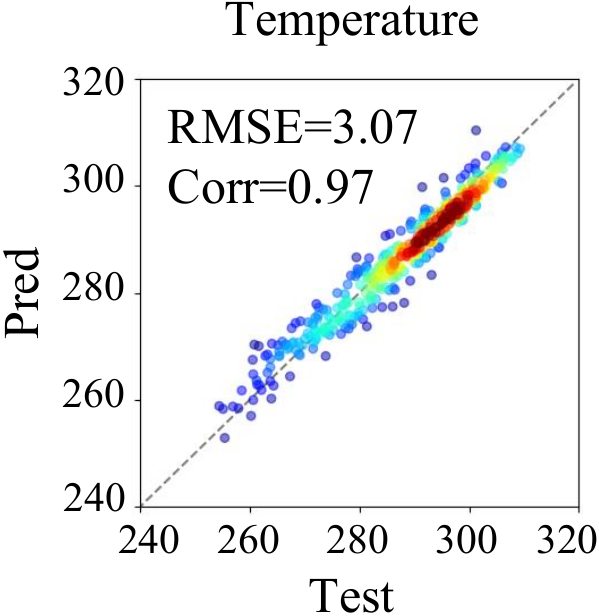}
}
\subfigure[$\mf_\mm$: forward approx.]{ \label{fig:c}{}
\includegraphics[width=0.79\columnwidth]{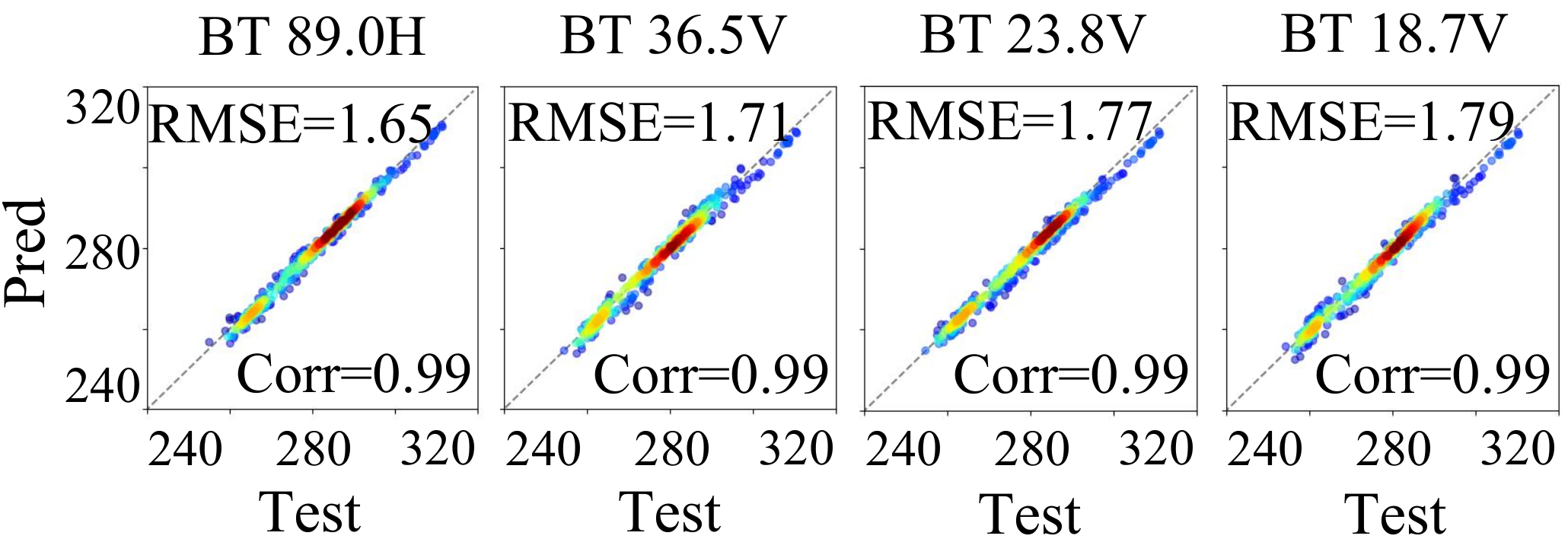}
}
\caption{Example approximation results on CMEM.
}
\label{fig:approx}
\end{figure}

\paragraph{Quality of inverse approximation.}
Fig. \ref{fig:approx} shows the results of inverse approximations for the physical model, where the inversion is necessary since the direction of simulation is often opposite to that of a prediction task. Here we use the CMEM model as an example, which simulates the process from the temperature to many different bands observed by the satellite (i.e., $\vy$ to $\vx$). Fig. \ref{fig:approx} (a) directly swaps the inputs and outputs of the physical model when training the network, whereas (b) uses the invertible network for the approximation. We can see that the regularization effects from the inversion can effectively reduce the RMSE and improve the approximation quality. For the original physical model direction, Fig. \ref{fig:approx} (c) includes examples of four approximated satellite bands using the invertible network to demonstrate that it works well in both directions.

\paragraph{Fairness results on AT1.}
The prediction performance and fairness results are shown in Tables \ref{tab:ew} to \ref{tab:state}, where each table corresponds to a different type of non-overlapping partitioning for training and testing.
We show the results of both FNN and LSTM in Table \ref{tab:ew} 
and keep FNN results in Table \ref{tab:t} 
and Table \ref{tab:state} 
as their trends are very similar.
All results are aggregated over 5 runs.
We use three metrics: RMSE, correlation coefficient (Corr.), and fairness (Eq. (\ref{eq:fair})).
\textbf{Geographic-region partitions: }
As shown in Table \ref{tab:ew}, the overall trend is that the two variants of SimFair consistently obtained the best fairness results for all three train-test splits.
Comparing different splits, SimFair methods have more consistent fairness results, whereas the other methods tend to perform better for the East-West split but worse for the other two splits.
It is interesting to note that the prediction performance (RMSE) of SimFair also tends to be much better than the other baseline approaches.
This is potentially due to the complimentary regularization effects brought by the deeper integration between the deep network and the simulation model using the dual-fairness consistency. 
\textbf{Temperature-zone and state-based partitions: }
Comparison results in Table \ref{tab:t} are similar to the previous partitioning, where SimFair continues to show the best performances in both fairness and prediction quality. 
It is worth noting that the performance is better in cold-to-hot than hot-to-cold scenarios.
The reason may be that temperature in colder regions is more stable and contains a narrower distribution, whereas it becomes more dynamic in hotter regions.
Table \ref{tab:state} demonstrates that SimFair is able to obtain fairer results in more local regions with a smaller amount of training data.

\begin{figure}[!t]
\centering
\includegraphics[width=0.86\columnwidth]{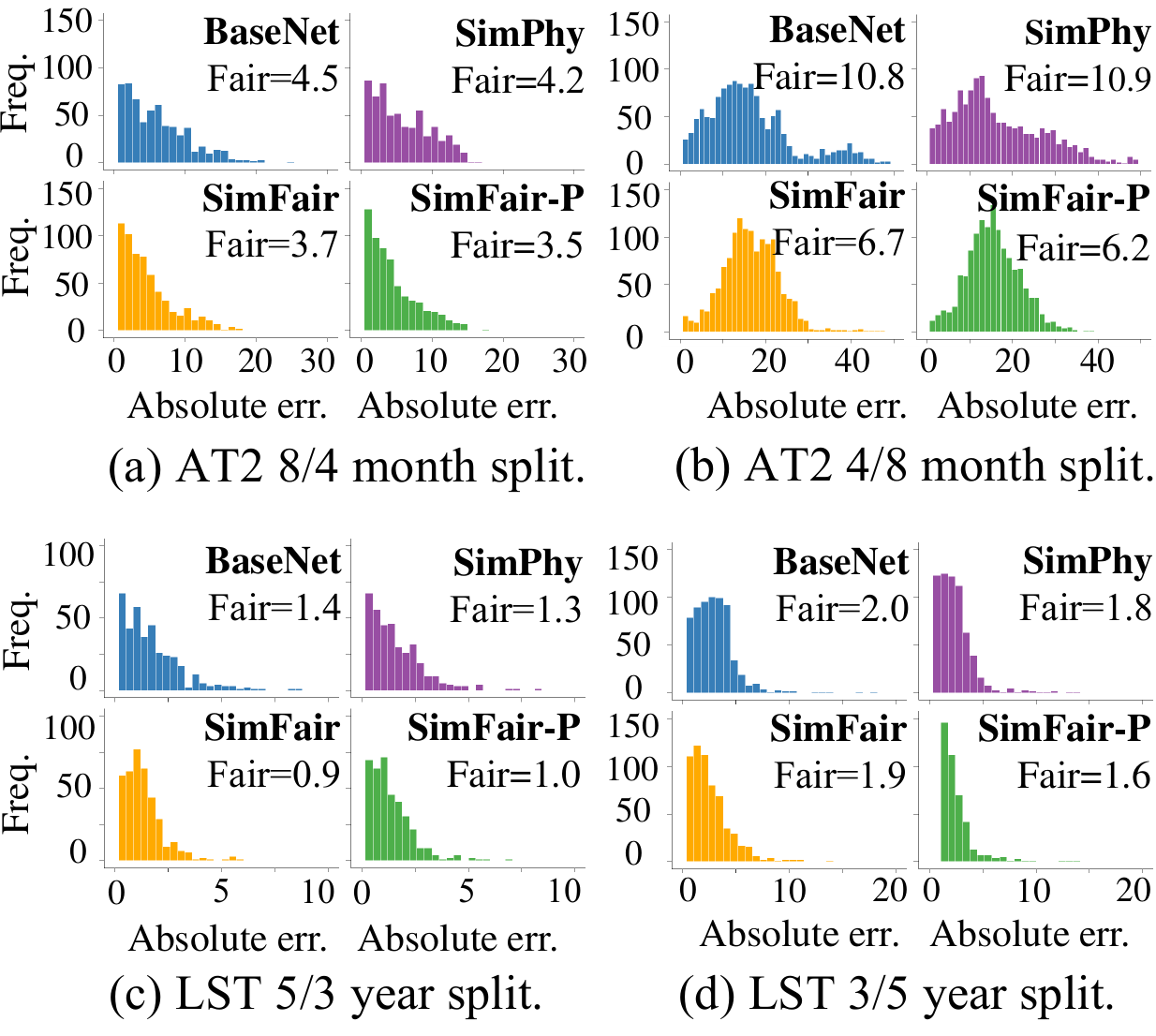}
\caption{Distributions of absolute errors in AT2 \& LST}
\label{fig:at2}
\end{figure}

\paragraph{Fairness results on AT2 and LST.}
We include additional results to see how well the methods can attach location-based fairness to the same set of locations over different periods.
Specifically, Fig. \ref{fig:at2} shows the absolute error distributions for the AT2 and LST datasets under various time splits for training and testing: (a) 8 and 4 months; (b) 4 and 8 months; (c) 5 and 3 years; and (d) 3 and 5 years.
For AT2, SimFair methods can reduce the variation of the prediction performance, and the 8/4-month split is easier for the methods.
Compared to the spatial tasks of AT1 in Table \ref{tab:ew}, the task here is overall easier based on the performance, as at least the groups (i.e., locations) used in fairness evaluation remain the same.
For LST, while BaseNet already performs well, SimFair methods are still able to further improve fairness scores.
\textbf{AT2/LST: Effects of physics models.}
For the results of the two physics-based models, CMEM for AT2 and MODTRAN for LST, SimFair methods perform well with both, showing that the general framework can potentially fit different types of simulations. 
Comparing CMEM and MODTRAN, the level of improvement is similar.

\section{Conclusions}
We proposed a SimFair framework to integrate physical simulation models into fairness-aware learning with inverse physical approximations, a dual-fairness consistency module, and physical constraints to promote fairer solutions. Our results on various simulation models and real datasets show SimFair can effectively improve fairness while keeping a similar (and sometimes better due to potential regularization effects) global performance as the baseline methods. Our future work will expand this to broader application domains and more knowledge- or rule-based simulation models.

\section*{Acknowledgments}
This material is based upon work supported by the National Science Foundation under Grant No. 2105133, 2126474 and 2147195; NASA under Grant No. 80NSSC22K1164 and 80NSSC21K0314; USGS under Grant No. G21AC10207; Google's AI for Social Good Impact Scholars program; the DRI award and the Zaratan supercomputing cluster at the University of Maryland; and Pitt Momentum Funds award and CRC at the University of Pittsburgh.

\bibliography{aaai24}

\clearpage

\begin{figure*}[!hbp]
\vspace{5pt}
\centering
\subfigure[West-East]{ \label{fig:a}{}
\includegraphics[width=0.60\columnwidth]{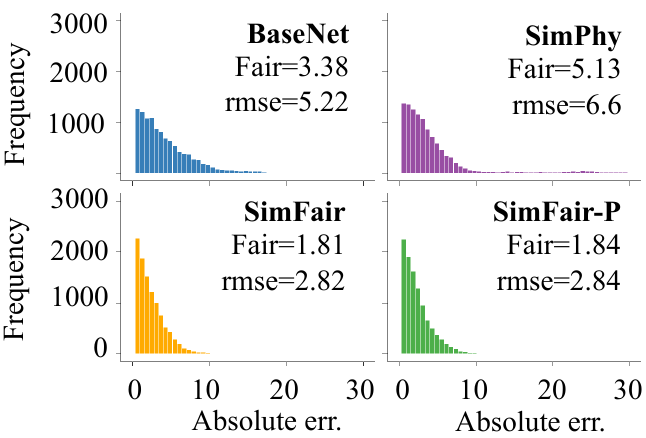}
}
\subfigure[East-West]{ \label{fig:b}{}
\includegraphics[width=0.60\columnwidth]{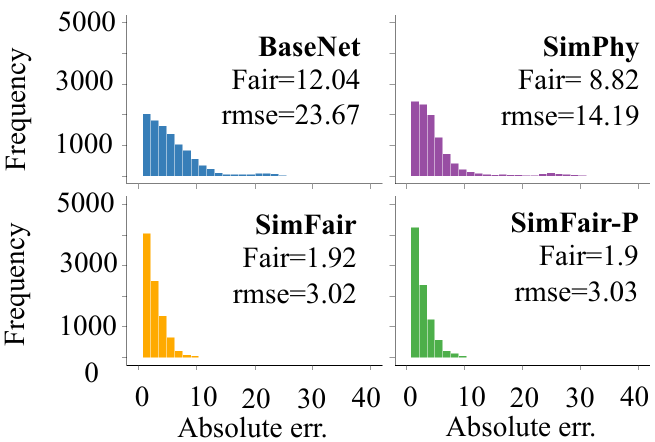}
}
\subfigure[East-Alaska]{ \label{fig:c}{}
\includegraphics[width=0.60\columnwidth]{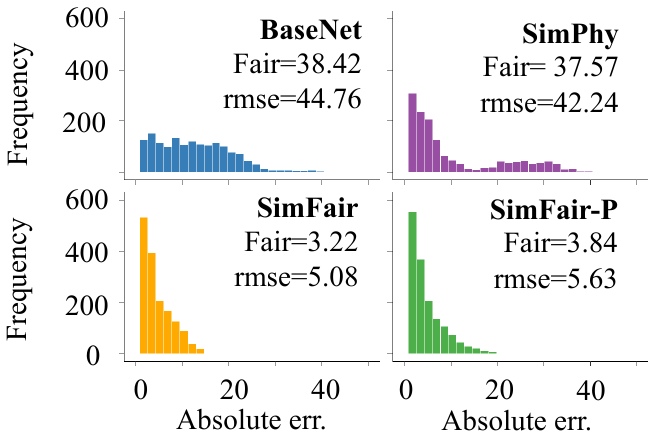}
}
\vspace{-12.4pt}
\caption{AT1: Distributions of absolute errors for geographic regions in Fig. \ref{fig:data}(a).
}
\label{fig:err_ew}
\end{figure*}

\begin{figure*}[bp!]
\centering
\subfigure[Hot-Cold]{ \label{fig:a}{}
\includegraphics[width=0.60\columnwidth]{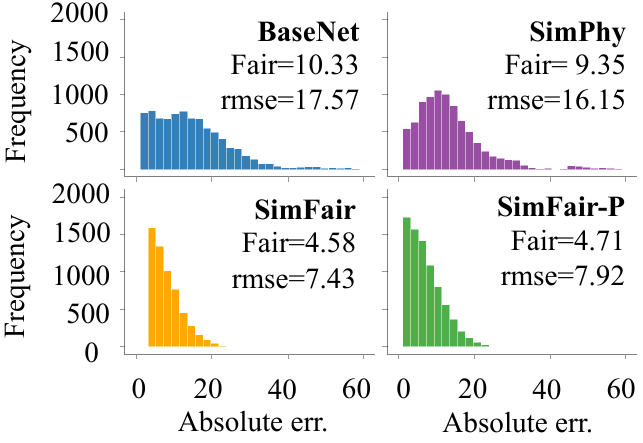}
}
\subfigure[Cold-Hot]{ \label{fig:b}{}
\includegraphics[width=0.60\columnwidth]{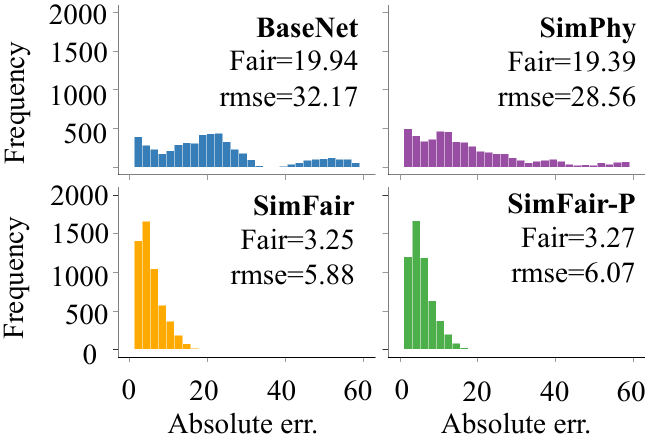}
}
\subfigure[Hot-Warm]{ \label{fig:c}{}
\includegraphics[width=0.60\columnwidth]{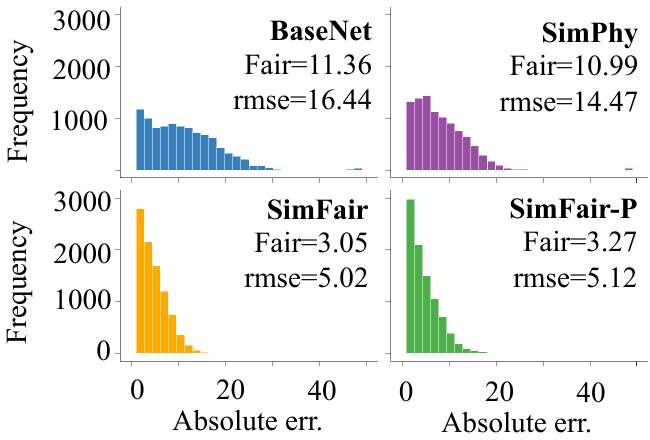}
}
\vspace{-12.4pt}
\caption{AT1: Distributions of absolute errors for temperature zones in Fig. \ref{fig:data}(b).
}
\label{fig:err_t}
\end{figure*}

\begin{figure*}[bp!]
\centering
\subfigure[Train-Test1]{ \label{fig:a}{}
\includegraphics[width=0.60\columnwidth]{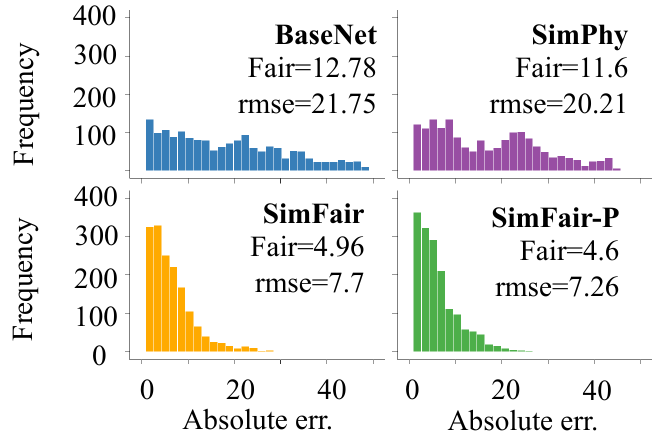}
}
\subfigure[Train-Test2]{ \label{fig:b}{}
\includegraphics[width=0.60\columnwidth]{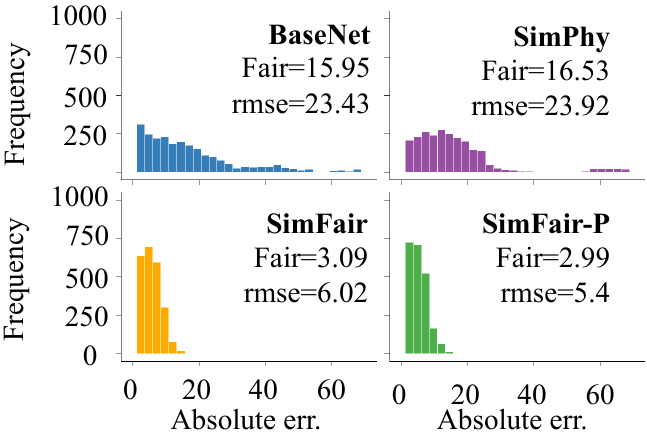}
}
\subfigure[Train-Test3]{ \label{fig:c}{}
\includegraphics[width=0.60\columnwidth]{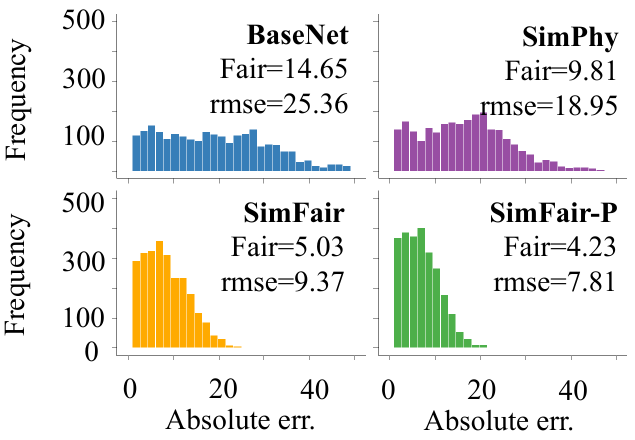}
}
\vspace{-12.4pt}
\caption{AT1: Distributions of absolute errors for random state groups in Fig. \ref{fig:data}(c).
}
\label{fig:err_states}
\end{figure*}

\section{Appendix}

\subsection{Implementation Details}
\paragraph{Fully-connected neural network (FNN).} FNN used three fully connected layers with 256 neurons and ReLU activation functions, ending with one output layer with 1 neuron for the predicted temperature. The batch size was 32, and all models were trained through 50 epochs. The optimizer was Adam, with an initial learning rate of $10^{-2}$, 150 as decay steps, and 0.96 as the decay rate of a scheduled exponential decay.

\paragraph{Long-short-term-memory (LSTM).} We used three bi-directional LSTM layers with 256 neurons and sigmoid activation functions. Between the output layer and LSTM layers, we used two fully-connected layers with 1024 and 128 neurons, respectively. The activation functions for the FNN are ReLU. The batch size, training epochs, optimizer, as well as learning rate share the same setting with the FNN networks.

\paragraph{Invertible network.} The invertible network has a chain of 7 bijectors where each bijector has 256 hidden neurons. The output dimension (e.g., satellite bands) equals the input dimensions (e.g., surface and atmospheric conditions). The batch size was 8, and training epochs are 50. The learning rate and optimizer are the same as the abovementioned two models.

\subsection{Additional Results}

First, we present additional visualizations of the error distributions of our results in Tables \ref{tab:ew},\ref{tab:t},\ref{tab:state} from the main text, as shown in Fig. \ref{fig:err_ew} to \ref{fig:err_states}.
Similarly, here a narrower distribution means a model has less variation of performance over locations, which is preferred for location-based fairness as defined in Eq. (\ref{eq:fair}). 
As we can see, the figures exhibit similar patterns as those from the main paper, and SimFair is able to improve the fairness in different scenarios.

Finally, Tables \ref{tab:lstm_tzone} to \ref{tab:lstm_state} show the LSTM-based results for the other two space partitionings for AT1, where Table \ref{tab:lstm_tzone} shows the results for the temperature-zone-based splits and Table \ref{tab:lstm_state} for the random-state-based version.
Similar to the FNN results (Tables \ref{tab:t} and \ref{tab:state}) in the main text, our proposed approaches, SimFair and SimFair-P, outperform the baseline models in most of the scenarios. 
While in some scenarios the prediction performance of some baselines are similar to or slightly better than the SimFair approaches, we can see our methods still maintain the best fairness scores in such cases, which is the main focus of the paper. For example, in the "Train-Test1" split in Table \ref{tab:lstm_state}, the RMSEs for BaseNet and SimFair are 10.13 and 10.49, respectively, and the fairness score (the lower the better) are 4.3 and 3.58, respectively.

\begin{table*}
\caption{AT1: Fairness results on temperature prediction (split by temperature zones in Fig. \ref{fig:data}(b)).
\vspace{-6pt}
}
\footnotesize
\centering
\begin{tabular}{|c|l||ccc||ccc||ccc|}
\hline
& & \multicolumn{3}{c||}{Train-Test: Hot-Cold}
& \multicolumn{3}{c||}{Train-Test: Cold-Hot}
& \multicolumn{3}{c|}{Train-Test: Hot-Warm}
\\\hline
& Model     & RMSE & Corr. & Fairness              & RMSE  & Corr. & Fairness              & RMSE  & Corr. & Fairness              \\
\hline
\multirow{8}{*}{\rotatebox[origin=c]{90}{\footnotesize LSTM}}

& BaseNet & 14.69 & 0.89 & 5.72 (±0.45)& 13.97 & 0.86 & 4.57 (±0.47)& 8.73 & 0.82 & 4.27 (±0.53) \\
& Sim & 13.78 & 0.86 & 5.8 (±0.72)& 14.78 & 0.88 & 4.36 (±0.14)& 9.92 & 0.9 & 4.36 (±0.96) \\
& SimPhy & 14.58 & 0.9 & 5.47 (±0.35)& 14.34 & 0.88 & 4.46 (±0.19)& 8.92 & 0.84 & 4.75 (±0.84) \\
& RegFair & 14.35 & 0.84 & 6.63 (±1.28)& 14.35 & 0.86 & 4.68 (±0.31)& 9.28 & 0.91 & 4.07 (±0.33) \\
& Self-Reg & 14.51 & 0.86 & 6.3 (±0.76)& 13.89 & 0.86 & 4.54 (±0.51)& 8.64 & 0.89 & 4.54 (±0.39) \\
\cline{2-11}
& SimFair & 14.65 & 0.91 & 5.42 (±0.52)& 13.66 & 0.9 & 3.78 (±0.37)& 7.24 & 0.92 & \textbf{3.88 (±0.2)} \\
& SimFair-P & 14.9 & 0.91 & \textbf{5.04 (±0.39)} & 11.69 & 0.92 & \textbf{3.34 (±0.34)} & 9.39 & 0.92 & 3.91 (±0.36) \\

\hline
\end{tabular}
\label{tab:lstm_tzone}
\end{table*}

\begin{table*}
\caption{AT1: Fairness results on temperature prediction (split by random state groups in Fig. \ref{fig:data}(c)). 
\vspace{-6pt}
}
\footnotesize
\centering
\begin{tabular}{|c|l||ccc||ccc||ccc|}
\hline
& & \multicolumn{3}{c||}{Train-Test: Train-Test1}
& \multicolumn{3}{c||}{Train-Test: Train-Test2}
& \multicolumn{3}{c|}{Train-Test: Train-Test3}
\\\hline
& Model     & RMSE & Corr. & Fairness              & RMSE  & Corr. & Fairness              & RMSE  & Corr. & Fairness              \\
\hline
\multirow{8}{*}{\rotatebox[origin=c]{90}{\footnotesize LSTM}}
& BaseNet & 10.13 & 0.82 & 4.3 (±0.89)& 4.34 & 0.92 & 2.8 (±0.11)& 7.5 & 0.92 & 4.44 (±0.28) \\
& Sim & 10.27 & 0.81 & 4.99 (±1.83)& 5.16 & 0.93 & 3.16 (±0.23)& 8.32 & 0.9 & 4.75 (±0.34) \\
& SimPhy & 7.34 & 0.88 & 3.98 (±0.44)& 5.51 & 0.92 & 3.38 (±0.43)& 9.3 & 0.94 & 4.93 (±0.21) \\
& RegFair & 9.7 & 0.86 & 4.36 (±1.32)& 5.73 & 0.86 & 4.29 (±2.56)& 6.7 & 0.9 & 4.7 (±1.24) \\
& Self-Reg & 13.56 & 0.65 & 9.22 (±7.72)& 5.81 & 0.88 & 3.44 (±0.79)& 6.78 & 0.91 & 4.4 (±0.16) \\
\cline{2-11}
& SimFair & 10.49 & 0.9 &  \textbf{3.58 (±0.32)}& 4.24 & 0.94 & 2.68 (±0.27)& 7.08 & 0.95 &  \textbf{4.24 (±0.26)} \\
& SimFair-P & 8.66 & 0.9 & 3.59 (±0.6)& 4.25 & 0.93 &  \textbf{2.64 (±0.33)}& 6.68 & 0.94 & 4.29 (±0.21) \\
\hline
\end{tabular}
\label{tab:lstm_state}
\end{table*}

\end{document}